\documentclass[letterpaper, 10 pt, conference]{ieeeconf}  

\IEEEoverridecommandlockouts                              

\usepackage{graphicx} 
\usepackage{epsfig} 
\usepackage{mathptmx} 
\usepackage{times} 
\usepackage{amsmath} 
\usepackage{amssymb}  
\usepackage{multirow}
\usepackage{url}
\usepackage{bm}
\usepackage{booktabs}
\usepackage{caption}
\usepackage[table]{xcolor}
\usepackage{array}
\usepackage{cite}

\newcommand{\tabsetup}{%
  \footnotesize
  \setlength{\tabcolsep}{4.5pt}%
  \renewcommand{\arraystretch}{1.18}%
}
\definecolor{ourrow}{gray}{0.92}
\newcolumntype{L}[1]{>{\raggedright\arraybackslash}p{#1}}

\begin{document}

\title{\LARGE \bf
ULOHA: An Underwater Bimanual Robot System for Robot Learning
}

  \author{Masato Kobayashi$^{1,2\dag*}$, Takeru Tsunoori$^{1\dag}$ 
  \thanks{
  ${\dag}$ Equal Contribution,
  $^{1}$ The University of Osaka, $^{2}$ Kobe University, * corresponding author: kobayashi.masato.cmc@osaka-u.ac.jp}
  }

\IEEEaftertitletext{%
    \begin{center}
      \includegraphics[width=\textwidth]{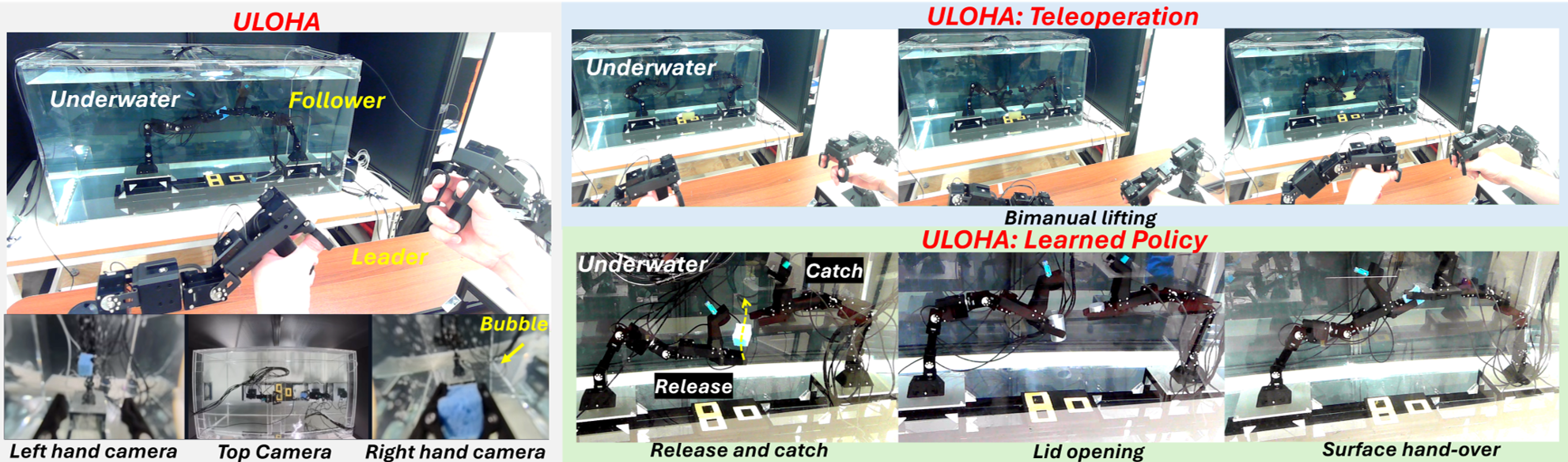}
      \captionof{figure}{ULOHA, an underwater bimanual robot learning platform.
        Left: two dry leader arms controlling two underwater follower arms,
        with example views from the left hand camera, top camera, and right
        hand camera below (left to right). Upper right: successive stages of
        bimanual lifting through leader--follower teleoperation, shown from
        left to right. Lower right: learned-policy execution of release and
        catch, lid opening, and surface hand-over (left to right).}
      \label{fig:teaser}
      \vspace{4mm}
    \end{center}
}
\maketitle

\thispagestyle{empty}
\pagestyle{empty}


\begin{abstract}

Underwater visuomotor policy learning has focused primarily on single
manipulators, while bimanual imitation learning has been studied largely in
air. We present ULOHA, an underwater bimanual robot learning platform that
combines custom-designed leader--follower hardware with software extensions
to LeRobot, integrating teleoperation, multi-view sensing, demonstration
collection, policy training, and autonomous deployment. Real-robot experiments
demonstrate a range of coordinated underwater bimanual behaviors, including
inter-arm transfer, shared-object manipulation, and buoyancy-driven
interception. We evaluate ACT, Diffusion Policy, and the vision--language--action
model SmolVLA on the platform. We investigate how learning methods
and execution strategies developed for manipulation in air perform underwater,
examining bubble disturbances, buoyancy-driven object motion, action-execution
horizons, and real-time chunking. A separate single-arm study examines
policy transfer between air and water and shows that demonstrations spanning
both media support execution in both under the tested conditions. ULOHA
provides a unified experimental platform for studying underwater bimanual robot learning
under the coupled perceptual and physical effects of underwater environments.
Additional material: \url{https://mertcookimg.github.io/uloha/}
\end{abstract}

\section{INTRODUCTION}

Platforms such as ALOHA and ALOHA 2 have made bimanual robot learning in air
more accessible by providing low-cost hardware and interfaces for collecting
demonstrations~\cite{zhao2023aloha,aloha2}. Such platforms provide an
experimental foundation for learning manipulation skills from
demonstrations~\cite{ravichandar2020lfd} and evaluating the resulting policies.
Comparable integrated platforms for underwater bimanual robot learning
remain less explored. Extending this experimental foundation to water would
enable studies of inter-arm transfer and shared-object manipulation under
additional physical and perceptual challenges. Buoyancy can move an object
after a gripper releases it, while bubbles can obscure the grippers and
object. Attenuation and scattering also alter underwater visual
observations~\cite{akkaynak2018model, akkaynak2019seathru}, motivating a platform for studying
learned coordination under these conditions.

Existing research provides important components of this foundation.
Underwater bimanual manipulation has been demonstrated through haptic
teleoperation~\cite{khatib2016oceanone} and online inverse-kinematics
learning~\cite{shen2024underwaterbimanual}. Recent underwater visuomotor learning
systems, including AquaBot~\cite{liu2025aquabot},
Bi-AQUA~\cite{tsunoori2025biaqua}, and UMI-Underwater~\cite{li2026umiuw},
have primarily addressed single manipulators. Together, these advances
motivate a platform that connects underwater bimanual demonstration collection
with policy training and autonomous execution, allowing existing learning
methods to be evaluated on coordinated tasks in real water.

We present ULOHA, an experimental platform for underwater bimanual robot
learning (Fig.~\ref{fig:teaser}). Two dry leader arms control two underwater
followers, with three cameras providing observations for demonstration
collection and policy execution. Waterproof cable connectors constrain the
followers' geometry and joint travel. Our custom structures accommodate these
constraints while preserving corresponding leader--follower joint layouts.
Together with ULOHA-specific extensions to LeRobot\cite{cadene2026lerobot}, the system supports
demonstration collection, policy training, and autonomous deployment on the
same hardware. The contribution centers on this integrated experimental
platform and its real-robot evaluation. We train or fine-tune established
ACT, Diffusion Policy, and SmolVLA models on underwater demonstrations.
Our contributions are:

\begin{itemize}
  \item \textbf{An integrated underwater bimanual robot learning platform}
        combining custom leader--follower hardware adapted to waterproof
        cabling, multi-view sensing, and a LeRobot-based pipeline for
        demonstration collection, policy training, and deployment.
  \item \textbf{A real-robot evaluation using established learning methods} across nine
        underwater tasks involving diverse forms of coordination, using ACT as the
        primary policy and evaluating Diffusion Policy and SmolVLA on selected tasks.
  \item \textbf{An empirical study of underwater policy deployment},
        examining bubble disturbances and execution strategies for
        buoyancy-driven interception, including action-execution horizons and
        real-time chunking. A separate single-arm ACT study evaluates how
        training-medium coverage affects air--water transfer.
\end{itemize}
To facilitate reproduction and extension, we will release the hardware designs and software as open source.

\section{RELATED WORK}

\subsection{Bimanual teleoperation and imitation learning}

ALOHA and ALOHA 2 enable low-cost bimanual demonstration collection by
mapping two leader arms to corresponding followers~\cite{zhao2023aloha,aloha2}.
Mobile ALOHA extends this approach to whole-body mobile manipulation~\cite{fu2024mobile},
while UMI collects transferable demonstrations using handheld grippers~\cite{chi2024umi}.
GELLO uses kinematically matched controllers across several arm platforms~\cite{wu2024gello}.
SO-101 provides accessible leader--follower hardware~\cite{huggingface_so101},
while OpenArm offers an open-source bimanual arm platform~\cite{enactic_openarm}.
MEVION uses unilateral leader--follower control for powerful, high-speed
bimanual manipulation~\cite{kawaharazuka2026mevion}.
ALPHA-$\alpha$ incorporates bilateral control to collect position
and force information for imitation learning~\cite{kobayashi2025alpha}.

Learning methods also differ in their action representations and
execution strategies. ACT predicts action
chunks~\cite{zhao2023aloha}; Diffusion Policy represents action
sequences through a conditional diffusion
process~\cite{chi2023diffusion}; and SmolVLA combines a pretrained
vision--language backbone with an action expert~\cite{shukor2025smolvla}.
RDT-1B develops a foundation model for bimanual
manipulation~\cite{liu2024rdt}, while PerAct$^2$ introduces a diverse
bimanual benchmark and a language-conditioned behavior-cloning
method~\cite{grotz2024peract2benchmarkinglearningrobotic}. Real-time
chunking addresses inference delays when executing action-chunking
flow policies~\cite{black2025rtc}.

These systems and methods primarily address manipulation in air.
ULOHA extends this line of work to demonstration collection
and autonomous policy execution for underwater bimanual manipulation.

\subsection{Underwater manipulation and learning}
\label{sec:classical}

Underwater manipulators span diverse mechanical and control designs~\cite{sivcev2018review}.
Underwater bimanual manipulation predates ULOHA. Ocean One combines
haptic teleoperation with whole-body control and has recovered artefacts at
sea~\cite{khatib2016oceanone,stuart2017oceanonehands}.
Sitler et al.~\cite{10597507} present an open-source framework for
teleoperating an underwater vehicle and two manipulators using two low-cost
haptic input devices. They demonstrate vehicle control and coordinated
bimanual grasping in a physics-based simulation.
Shen et al. learn inverse kinematics online for a bimanual soft
manipulator~\cite{shen2024underwaterbimanual}.
LURE~\cite{allen2025lure} combines Laplacian trajectory editing, optimal
control, and symbolic planning to adapt a demonstrated trajectory for
execution with a single underwater manipulator.

Recent underwater visuomotor policy learning has primarily focused on a
single arm or gripper. AquaBot learns ROV-gripper manipulation
from human demonstrations and improves execution speed through
self-learning~\cite{liu2025aquabot}. Bi-AQUA adds lighting-aware conditioning to
bilateral imitation learning on a three-joint arm with a
gripper~\cite{tsunoori2025biaqua}. UMI-Underwater transfers a depth-based grasp-affordance model
learned from on-land demonstrations and trains an affordance-conditioned
diffusion policy using autonomously collected underwater
demonstrations~\cite{li2026umiuw}.
USIM and U0 extend underwater VLA learning in simulation~\cite{gu2025usim}.
Table~\ref{tab:related} compares selected platforms.

ULOHA provides an integrated platform for collecting bimanual demonstrations,
training established policy models, and deploying the learned policies on
physical robots in water. It provides a
real-water setting for evaluating established learning methods
and action-execution strategies, including RTC, on coordinated
bimanual tasks. A separate single-arm ACT study examines direct
air--water policy transfer and mixed-medium training.

\begin{table}[t]
\caption{Comparison of on-land and underwater manipulation systems.}
\label{tab:related}
\centering
\tabsetup
\setlength{\tabcolsep}{2pt}
\begin{tabular}{@{}l c c c c c@{}}
\toprule
System & Medium & Manip. & \shortstack{Act.\\joints} &
\shortstack{Learned\\visuomotor\\policy} &
\shortstack{Bimanual\\learned\\execution} \\
\midrule
\shortstack[l]{ALOHA / ALOHA 2\\\cite{zhao2023aloha,aloha2}} & air & 2 & $7\times2$ & $\checkmark$ & $\checkmark$ \\
\addlinespace[1pt]
AquaBot~\cite{liu2025aquabot} & water & 1 & $1^{*}$ & $\checkmark$ & --- \\
Bi-AQUA~\cite{tsunoori2025biaqua} & water & 1 & 4 & $\checkmark$ & --- \\
UMI-Underwater~\cite{li2026umiuw} & water & 1 & 1 & $\checkmark$ & --- \\
\midrule
\rowcolor{ourrow}
\textbf{ULOHA (ours)} & \textbf{water}$^{\dagger}$ & \textbf{2} & $\bm{7\times2}$ & $\checkmark$ & $\checkmark$ \\
\bottomrule
\addlinespace[2pt]
\multicolumn{6}{@{}p{0.96\columnwidth}@{}}{\scriptsize
  $\checkmark$: demonstrated in the cited work; ---: not demonstrated.
  Manip. counts arms or standalone grippers. Joint counts include grippers
  and exclude vehicle motion and thrusters.
  $^{*}$AquaBot's entry counts one commanded gripper opening/closing function.
  $^{\dagger}$ULOHA's bimanual experiments are underwater; only the single-arm
  medium study uses air as well.}
\end{tabular}
\end{table}

\begin{figure*}[t]
  \centering
  \includegraphics[width=\textwidth]{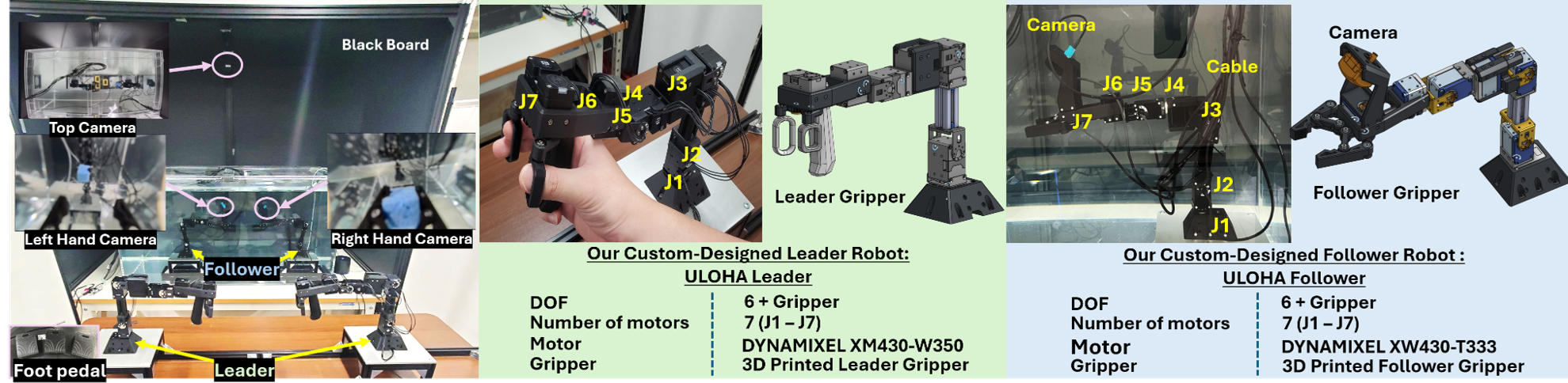}
  \caption{Custom-designed ULOHA hardware. Left: arrangement of
    the dry leaders, underwater followers, three cameras, and foot pedal.
    Center and right: photographs, CAD models, and specifications of the
    leader and follower arms, respectively. Joint labels J1--J6 identify
    the arm joints, and J7 identifies gripper actuation. The leaders use
    DYNAMIXEL XM430-W350 servos, and the followers use waterproof
    DYNAMIXEL XW430-T333 servos. Both arms use custom 3D-printed grippers.}
  \label{fig:hardware}
\end{figure*}

\begin{figure}[t]
  \centering
  \includegraphics[width=\linewidth]{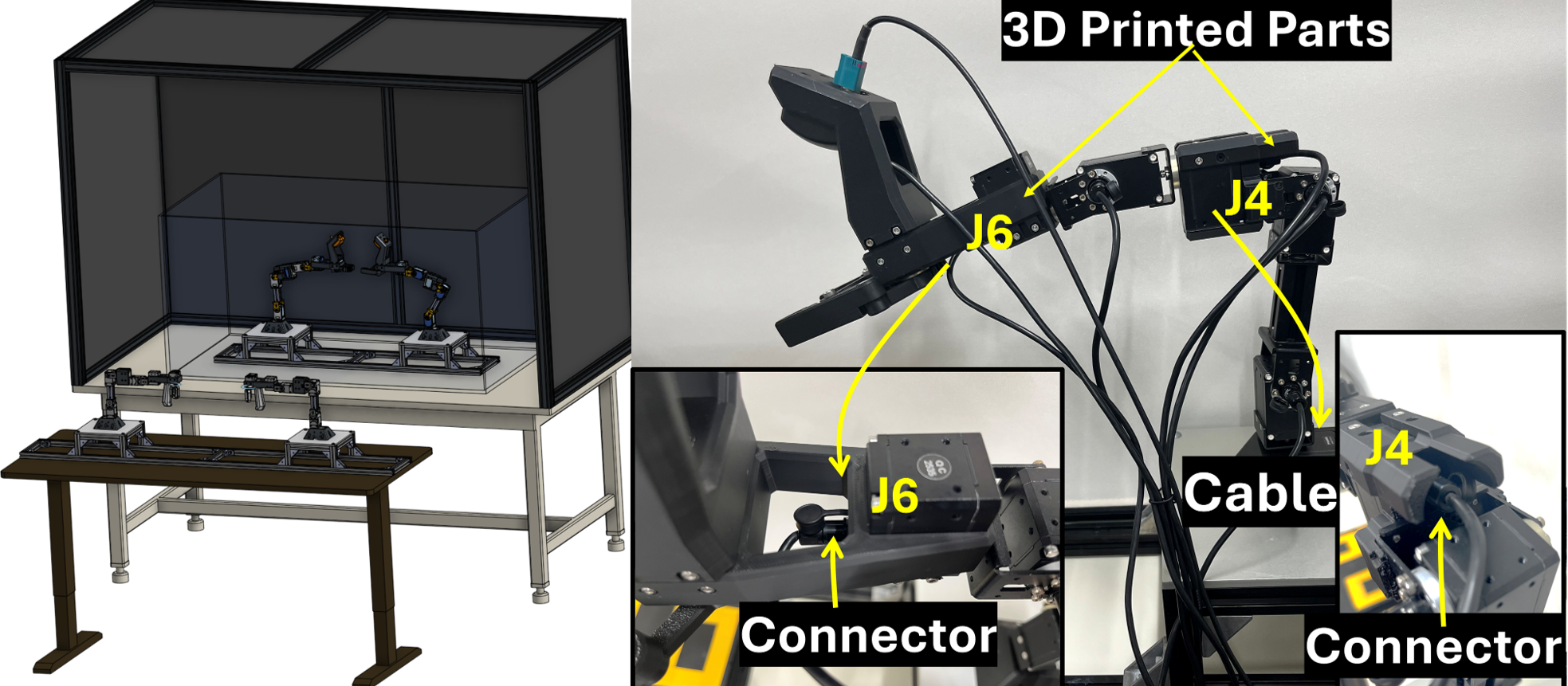}
  \caption{Experimental setup and follower connector accommodation.
    Left: CAD model of the dry leaders and tank-mounted followers within
    an enclosure made from aluminum frames and black panels. Right:
    follower arm with custom 3D-printed parts and individual waterproof
    cables. Insets show connector clearance at J6 (lower left of the
    right panel) and the connector region at J4 (lower right), where
    arm motion is mechanically limited to prevent interference.}
  \label{fig:tank}
\end{figure}
\section{ULOHA SYSTEM}
\label{sec:platform}

\subsection{System overview}

ULOHA combines hardware designed by the authors with a software stack that
extends LeRobot~\cite{cadene2026lerobot} through ULOHA-specific hardware
adapters and recording, training, and deployment configurations. The hardware
comprises two dry leader arms and two kinematically matched underwater
followers, with custom arm structures and grippers. During teleoperation, each follower tracks its corresponding leader while joint states, images, and language task descriptions are recorded; during autonomous execution, a learned policy supplies joint
targets to the follower controller.

The LeRobot-based stack stores language task descriptions with demonstrations and provides them alongside images and robot states to compatible policies, enabling vision--language--action (VLA) models such as SmolVLA to be deployed on the underwater hardware.
ULOHA builds on LeRobot to support the integration of different policy models.
In this study, we evaluate ACT, Diffusion Policy, and SmolVLA on selected tasks.
Hardware, perception, demonstration collection, and deployment are described
below.

\subsection{Leader--follower hardware}

We designed the four arms to retain corresponding joint layouts despite
different actuator and cabling requirements (Fig.~\ref{fig:hardware}). Each arm
has six arm joints and an actuated gripper. The leaders operate in air with
Dynamixel XM430-W350 servos, while the underwater followers use waterproof
XW430-T333 servos. The arms combine commercially available components,
custom 3D-printed parts, and aluminum framing; their bases are built from
aluminum frames. Both the leader and follower grippers are custom designed.
The corresponding joint arrangement supports joint-position teleoperation
from the dry leaders to the underwater followers.

The principal mechanical constraint is the followers' waterproof cabling.
Unlike the leaders, the followers cannot use direct motor-to-motor
daisy-chain connections in this configuration: each motor requires a separate
waterproof cable. We therefore designed the follower structures around the
motor-side cable connectors to avoid interference during joint motion
(Fig.~\ref{fig:tank}, right). At follower joint J4, a mechanical motion
limit prevents the arm structure from colliding with the connector. At J6,
the structure leaves space around the connector to avoid interference.
Thus, the leader and follower share a joint layout, but the follower's
mechanically permitted motion is constrained by its connector geometry.
These adaptations accommodate underwater wiring while retaining the joint
correspondence needed for collecting bimanual demonstrations from an operator
in air. They address a physical integration problem that replacing the
leaders' servos with waterproof models alone would leave unresolved.

The XW430-T333 servos allow immersion without a separate pressure housing per
joint. Their IP68 rating is specified for fresh water at a depth of at most
1\,m for 24\,h~\cite{robotisxw430}; underwater sessions are kept within this
envelope. Waterproof cables and connections must also be suitable for immersion
and correctly assembled~\cite{robotisxw430}.

Each arm uses a separate U2D2 bus; motor identifiers are specified in the
experiment configurations. The 14-dimensional follower state and action vectors
are ordered left-then-right.

The costs are approximately USD~2,600 per leader and
USD~7,000 per follower.

\begin{figure*}[t]
  \centering
  \includegraphics[width=\textwidth]{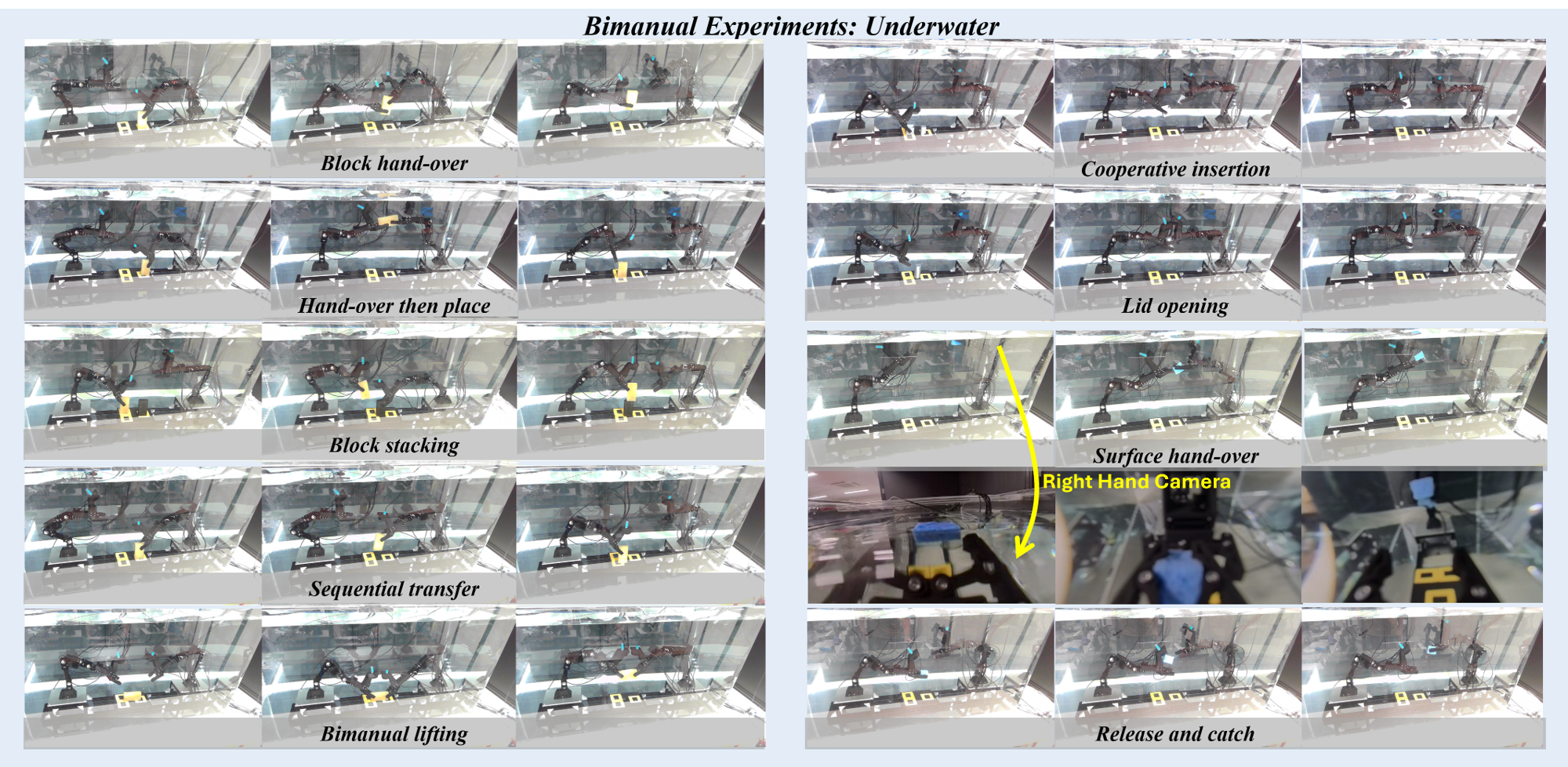}
  \caption{Autonomous ACT rollouts across nine underwater bimanual
    tasks. For each task, selected frames show the progression from left to
    right. The surface hand-over inset shows views from the right hand camera.
    Success rates are reported in Table~\ref{tab:bitasks}.}
  \label{fig:bitasks}
\end{figure*}
\subsection{Perception and data acquisition}

The tank measures $1.2\,\mathrm{m}$ in width, $0.6\,\mathrm{m}$ in depth, and
$0.6\,\mathrm{m}$ in height (Fig.~\ref{fig:tank}). An enclosure made from
aluminum frames and black plastic panels surrounds the workbench supporting
the tank. This enclosure is intended to reduce variation in the visual
background and ambient lighting across experimental environments, providing
more consistent conditions for camera observations.

Three TIER IV C1-120 GMSL2 cameras are attached through USB capture kits: one top camera
that sees both arms and the workspace, and one hand camera per arm. The kits
expose a single native mode, $1920\times1280$ UYVY, so frames are captured at
that resolution and resized to $480\times320$ --- preserving the 3:2 sensor aspect
ratio --- before AV1 encoding at 30\,fps. Observation keys are prefixed with the
arm's side and documented in the configurations.

\subsection{Teleoperation and imitation learning}

Demonstrations are collected by unilateral teleoperation with leader torque
disabled. LeRobot's calibration and normalization map corresponding joints
between independently calibrated leader and follower ranges, accommodating
their different travel limits.

At 30\,Hz, we record normalized leader targets, follower states, and three
camera streams.
During deployment, the policy consumes the
three images and 14-dimensional follower state and predicts normalized action
sequences. A configurable prefix is executed at a nominal 30\,Hz before
re-planning. The pipeline builds on LeRobot~\cite{cadene2026lerobot}, with
ULOHA hardware adapters and configuration-based recording, training, and rollout.

\section{EXPERIMENTAL EVALUATION}
\label{sec:eval}

\subsection{Experimental setup}
\label{sec:protocol}

The evaluation addresses four questions: which coordinated underwater tasks
can be learned from demonstrations; whether ACT, Diffusion Policy, and SmolVLA
can be deployed on the platform; how bubbles and action-execution strategies affect deployment;
and whether policies transfer between air and water. We first evaluate
bimanual learning and deployment, then use a separate single-arm ACT study to
examine training-medium coverage.

Throughout the experiments, white blocks are polyurethane blocks, black
blocks are rubber blocks, and the blue sponges are cellulose sponges.

\subsubsection{Underwater bimanual tasks}

The nine underwater tasks and their success states are defined below
(Fig.~\ref{fig:bitasks}); Table~\ref{tab:bitasks} summarizes the arms' roles.
\begin{itemize}
  \item \emph{Block hand-over.} The right arm transfers a block to the left
        without the block touching the tank floor; the left gripper holds it at completion.
  \item \emph{Hand-over then place.} After the same transfer without floor contact,
        the left arm places the block fully inside the target region at rest.
  \item \emph{Block stacking.} Each arm handles one block; the white
        polyurethane block must rest stably on the black rubber block.
  \item \emph{Sequential transfer.} The right arm completes the first leg of
        a block-transfer route, and the left completes the second after an
        intermediate placement.
  \item \emph{Bimanual lifting.} Both grippers jointly grasp and hold a block
        clear of the tank floor.
  \item \emph{Cooperative insertion.} The left arm holds a cup upright while
        the right releases a block inside it.
  \item \emph{Lid opening.} The right arm separates the lid while the
        container remains held by the left gripper.
  \item \emph{Surface hand-over.} The right arm lifts a floating blue
        cellulose sponge off the water surface and transfers it to the left.
  \item \emph{Release and catch.} The left arm releases a underwater blue
        cellulose sponge; the right catches it before it reaches the surface.
\end{itemize}

\begin{figure*}[t]
  \centering
  \includegraphics[width=\textwidth]{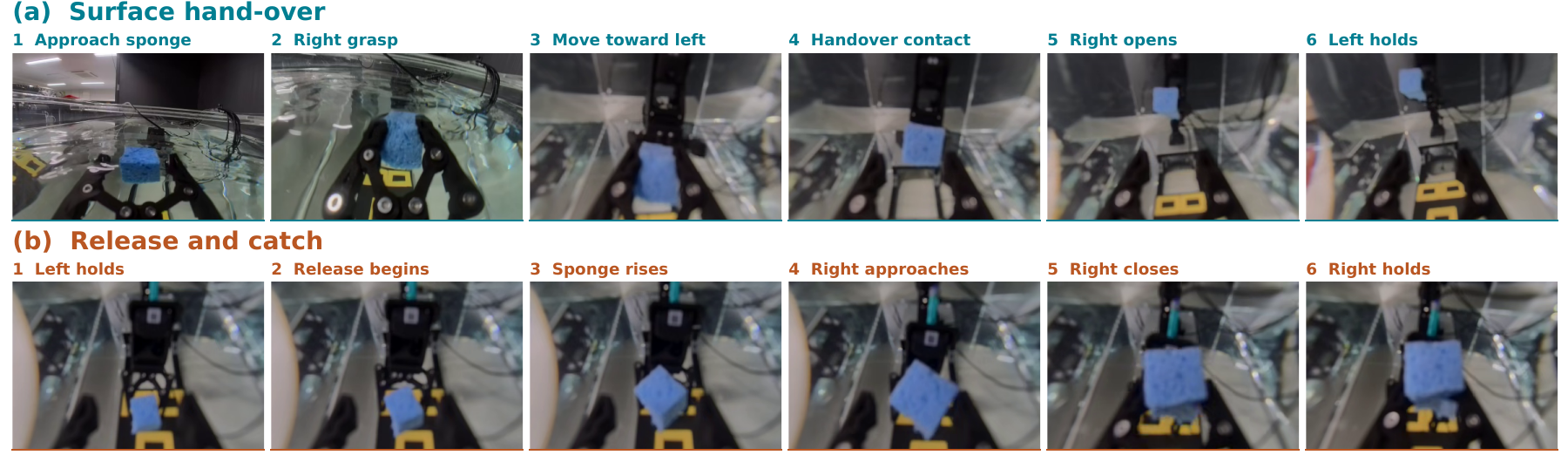}
  \caption{Views from the right hand camera during demonstrations of buoyant blue
  cellulose sponge manipulation.
  (a) Surface hand-over: the right gripper grasps a floating sponge and
  transfers it to the left gripper. (b) Release and catch: the left gripper
  releases the sponge, and the right gripper intercepts it as it rises.}
  \label{fig:wrist_sponges}
\end{figure*}

Fig.~\ref{fig:wrist_sponges} shows views from the right hand camera during teleoperated
 demonstrations of the two sponge tasks, illustrating grasp transfer and
 release followed by interception.

\subsubsection{Demonstration datasets}

Surface hand-over and release and catch use 50 demonstrations each; the other
seven bimanual tasks use ten each. Single-arm datasets contain ten episodes
per medium, with the mixed set combining five per medium. Single-arm data use
two cameras; bimanual data use three.

\subsubsection{Policy training and inference}

ACT and Diffusion Policy are trained for 50k steps with batch sizes 8 and
32, respectively. SmolVLA is trained for 200k steps with an effective batch
size of 32.
ACT is used in both unimanual and bimanual tasks,
whereas Diffusion Policy and SmolVLA are evaluated only on bimanual tasks.

The prediction horizon specifies the number of actions generated per query;
the execution horizon specifies how many are executed before re-planning.
Baseline prediction/execution horizons are 100/100 for ACT, 64/32 for Diffusion
Policy, and 50/50 for synchronous SmolVLA. The ACT horizon and SmolVLA RTC
studies modify execution as described in their respective sections.

ACT~\cite{zhao2023aloha} uses a single ResNet-18
backbone~\cite{he2016resnet} shared across camera views. Its transformer has
width 512 and eight heads, with dropout $0.1$, KL weight $10$, and temporal
ensembling disabled; backbone and transformer learning rates are $10^{-5}$.
Diffusion Policy~\cite{chi2023diffusion} uses two observation steps, random
cropping at ratio $0.95$, and an end-of-episode window guard of 31.
SmolVLA~\cite{shukor2025smolvla} is fine-tuned from the public base checkpoint
with the vision encoder unfrozen. Learning-rate decay horizons follow the
configured training step count.

\begin{table}[t]
\caption{ACT success on nine underwater bimanual tasks in clear
water using the 100-step execution baseline. Ten evaluation trials per task;
the final two tasks use 50 demonstrations each, and the others use ten.
The final two tasks involve buoyancy-driven sponge manipulation.}
\label{tab:bitasks}
\centering
\tabsetup
\setlength{\tabcolsep}{3.4pt}
\begin{tabular}{@{}L{0.27\columnwidth} L{0.38\columnwidth} r@{}}
\toprule
Task & Role & Success \\
\midrule
Block hand-over      & Pass and receive a block   & \textbf{10/10} \\
Hand-over then place  & Pass, receive, and place & 6/10 \\
Block stacking        & Align and stack two blocks     & 9/10 \\
Sequential transfer   & Complete successive route legs   & \textbf{10/10} \\
Bimanual lifting      & Grasp and lift jointly  & \textbf{10/10} \\
Cooperative insertion & Hold cup and insert block          & 9/10 \\
Lid opening           & Hold container and remove lid   & 7/10 \\
\midrule
Surface hand-over     & Grasp and transfer floating sponge & 4/10 \\
Release and catch     & Release and intercept rising sponge & 3/10 \\
\midrule
\rowcolor{ourrow}
\textbf{Total} & --- & \textbf{68/90} \\
\bottomrule
\addlinespace[2pt]
\multicolumn{3}{@{}p{\columnwidth}@{}}{\scriptsize
  Ten trials per task. The total sums successes across all nine tasks.}
\end{tabular}
\end{table}

\subsubsection{Evaluation protocol}

Each condition comprises ten trials, with a 60\,s limit for bimanual tasks
and 45\,s for single-arm tasks, including the unseen-object condition.
Success requires reaching the task-specific state within the limit and
retaining it at episode end. Each task uses a separately trained policy.
Bubble and execution-strategy comparisons reuse the relevant baseline
checkpoint without retraining. Baseline trials are reused where indicated
and counted only once: 90 baseline ACT trials across the nine bimanual tasks, 20 bubble trials, 20 additional block hand-over policy trials, 30 additional ACT execution-horizon trials,
20 SmolVLA release-and-catch trials, and 70 single-arm trials, totaling 250 trials.

\subsection{Underwater bimanual learning}
\label{sec:bimanual}

We assess the range of coordinated behaviors learned with ACT and the
deployment of Diffusion Policy and SmolVLA on the same platform.

\subsubsection{Nine underwater bimanual tasks}

We evaluate ACT across the nine tasks to characterize the range of
learned bimanual behaviors (Table~\ref{tab:bitasks}). Each task uses its own
checkpoint and the baseline 100-step execution horizon. Success ranges from
3/10 to 10/10, with a descriptive total of 68/90 across task-specific policies.
Both arms successfully participate in transfers, shared lifting, and
complementary manipulation roles. With the 100-step execution baseline,
the lowest rates occur on surface hand-over (4/10) and release and catch (3/10).

\begin{figure}[t]
\centering
\includegraphics[width=\columnwidth]{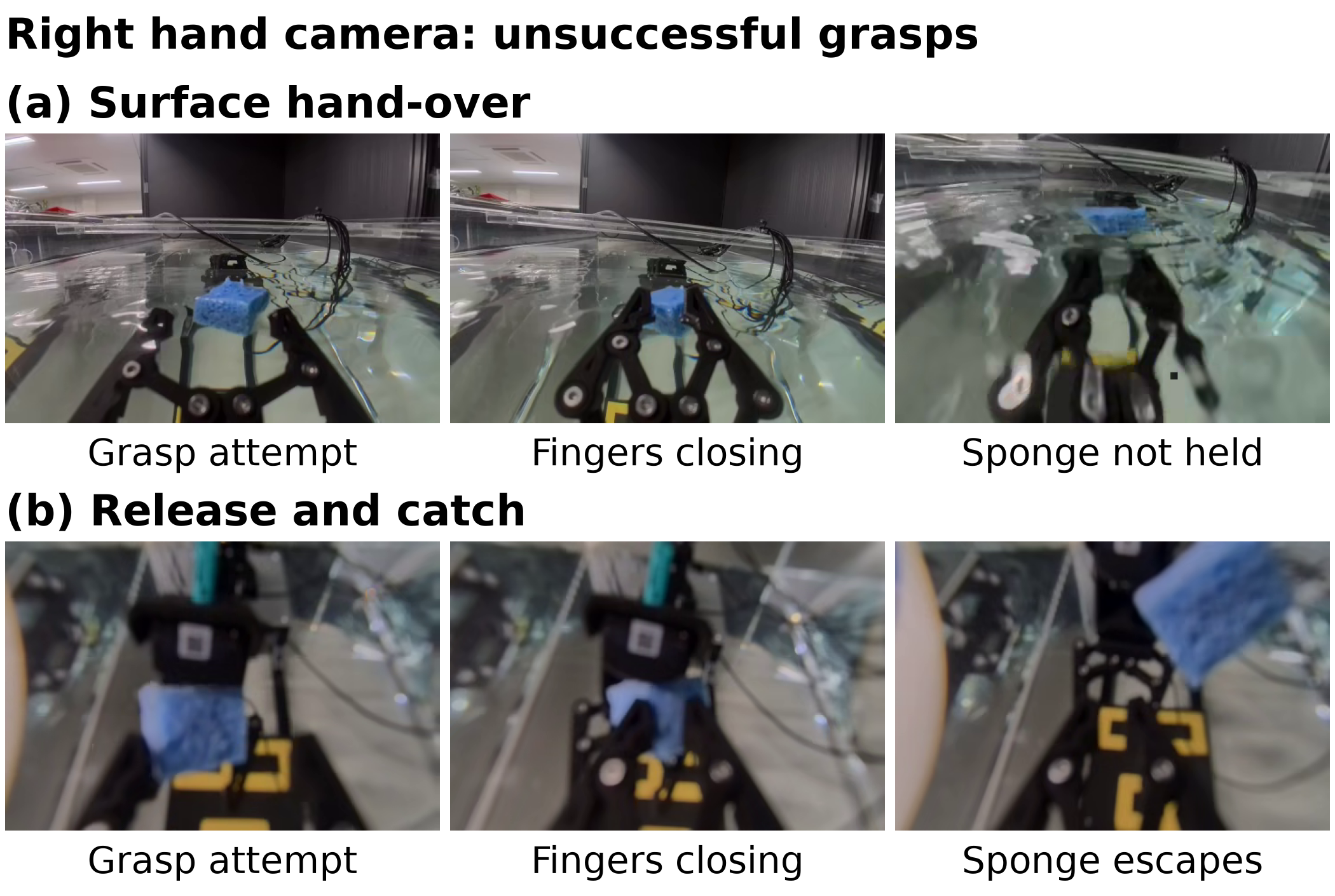}
\caption{Unsuccessful sponge grasp attempts viewed by the right hand camera.
(a) In surface hand-over, the sponge is between the fingers during closure
but remains at the water surface as the gripper withdraws.
(b) In release and catch, the sponge is not retained and escapes upward.}
\label{fig:sponge_failures}
\end{figure}

Both sponge tasks require a nearly fully open gripper, leaving little
clearance for positional error. Fig.~\ref{fig:sponge_failures} shows two
unsuccessful grasp attempts. In surface hand-over, the sponge is between the
fingers during closure but remains at the water surface as the gripper
withdraws. In release and catch, the sponge is not retained and escapes
upward. Release and
catch requires interception during ascent; initial grasping in surface
hand-over requires alignment with a sponge that can move horizontally on
the water surface. In the latter task, the right hand camera also crosses
from water into air, changing the visual conditions during approach.
We further examine execution-strategy sensitivity on release and catch
below (Section~\ref{sec:execution}).

\subsubsection{Evaluation with ACT, Diffusion Policy, and SmolVLA}
\label{sec:generality}

We evaluate ACT, Diffusion Policy, and SmolVLA on block hand-over using the same ten
demonstrations and robot initial pose. Each achieves 10/10. This demonstrates
that all three methods can be deployed on ULOHA for this task.

\subsection{Underwater deployment sensitivities}

We examine deployment under bubble disturbances and compare action-execution
strategies on release and catch. Each comparison holds the trained checkpoint
fixed to assess changes at deployment time.

\begin{figure}[t]
  \centering
  \includegraphics[width=\linewidth]{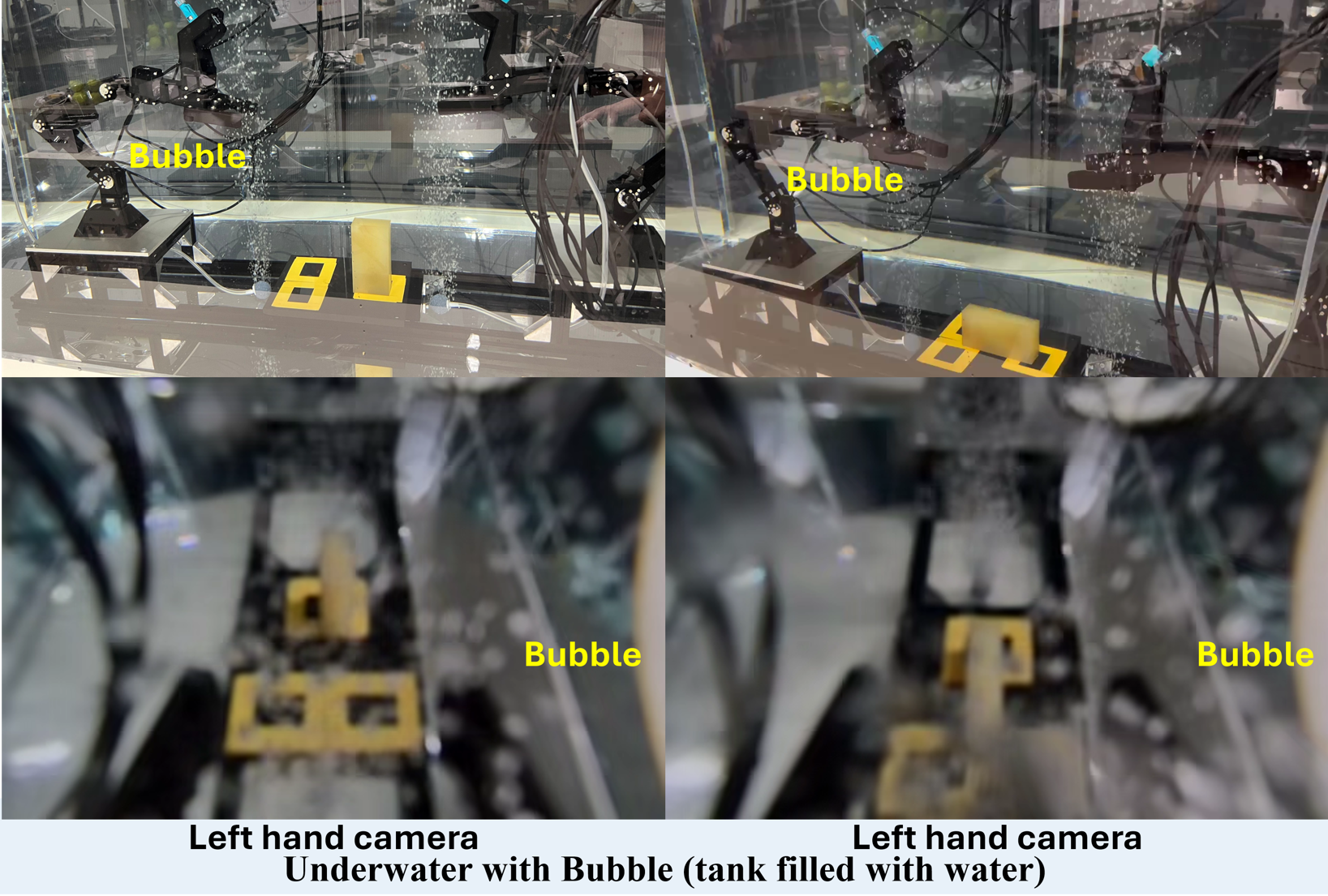}
  \caption{Bubble disturbances during sequential transfer (left column) and
    bimanual lifting (right column). Each column shows an external view of
    the underwater arms and bubble streams above and a corresponding left
    hand camera view below. ACT results with and without bubbles are reported in
    Table~\ref{tab:bubble}.}
  \label{fig:bubbles}
\end{figure}
\begin{table}[t]
\caption{ACT success with and without bubble disturbances.
The same checkpoints are evaluated without retraining.}
\label{tab:bubble}
\centering
\tabsetup
\small
\renewcommand{\arraystretch}{1.30}
\begin{tabular*}{\columnwidth}{@{\extracolsep{\fill}}L{0.36\columnwidth} c c@{}}
\toprule
\textbf{Task} & \textbf{Clear} & \textbf{Bubbles} \\
\midrule
Sequential transfer & \textbf{10/10} & 3/10 \\
Bimanual lifting    & \textbf{10/10} & \textbf{10/10} \\
\bottomrule
\end{tabular*}
\end{table}

\subsubsection{Performance under a bubble disturbance}
\label{sec:bubble}

We test whether policies trained in clear water remain effective under
bubble disturbances. Two aerators generate bubble streams through the
workspace (Fig.~\ref{fig:bubbles}); sequential transfer and bimanual lifting
are repeated with the same ACT checkpoints, robot initial pose, and success
criteria, without retraining. Sequential transfer drops from 10/10 to 3/10,
while bimanual lifting remains at 10/10 (Table~\ref{tab:bubble}).

Fig.~\ref{fig:bubbles} shows bubbles visible in the left hand camera views
for both tasks. Visual interference from bubbles may contribute to the
degradation in sequential transfer, but these example images do not
establish a difference in the duration of occlusion between tasks or
isolate its contribution to the success rates.

\begin{table}[t]
\caption{ACT execution horizon on release and catch. All settings use the same
checkpoint and 100-step prediction horizon.}
\label{tab:execution}
\centering
\tabsetup
\small
\renewcommand{\arraystretch}{1.30}
\begin{tabular*}{\columnwidth}{@{\extracolsep{\fill}}r c c@{}}
\toprule
\textbf{Executed steps} & \textbf{Nominal interval (s)} & \textbf{Success} \\
\midrule
100 (baseline) & 3.33 & 3/10 \\
50 & 1.67 & 4/10 \\
30 & 1.00 & \textbf{6/10} \\
15 & 0.50 & 3/10 \\
\bottomrule
\end{tabular*}
\end{table}
\subsubsection{Execution strategies on release and catch}
\label{sec:execution}

We study execution strategies for intercepting a rising sponge using ACT and
SmolVLA trained on the same 50 demonstrations. Each model's checkpoint is
held fixed across its execution conditions, without retraining. ACT's
comparison varies the number of actions executed before re-planning;
SmolVLA's comparison varies synchronous versus RTC execution.

\emph{ACT execution horizon.} The ACT checkpoint predicts 100 actions at each query. We execute prefixes of 100, 50, 30, or
15 actions, discard the remainder, and query the policy again; no retraining
is performed. At the nominal 30\,Hz control rate, these correspond to the
re-planning intervals in Table~\ref{tab:execution}.

The best observed result is 6/10 at 30 executed actions, compared with 3/10
for the 100-action baseline. Further shortening the prefix to 15 actions
returns success to 3/10. The relationship is therefore non-monotonic for
this checkpoint: more frequent re-planning alone does not ensure higher
success. The three additional settings contribute 30 trials; the baseline
row reuses the nine-task evaluation.

\label{sec:rtc}

\emph{SmolVLA real-time chunking.} We evaluate RTC~\cite{black2025rtc} as an asynchronous execution strategy
for SmolVLA on the same interception task. RTC generates a new action chunk
while the previous chunk is executing, conditioning on its remaining plan.
The execution-horizon parameter controls prefix
conditioning. Elapsed actions are skipped when the new chunk enters the
queue to account for inference delay.

Using the same checkpoint without additional training, RTC achieves
6/10 successes versus 4/10 for synchronous execution (Table~\ref{tab:rtc}).
Each strategy is evaluated in ten trials.
This demonstrates RTC deployment on a buoyancy-driven bimanual task.
The observed rate equals ACT's best
result at a 30-step execution horizon, without establishing that either
method outperforms the other.

\begin{table}[t]
\caption{SmolVLA execution strategies on release and catch.}
\label{tab:rtc}
\centering
\tabsetup
\small
\renewcommand{\arraystretch}{1.30}
\begin{tabular*}{\columnwidth}{@{\extracolsep{\fill}}l c@{}}
\toprule
\textbf{Execution strategy} & \textbf{Success} \\
\midrule
Synchronous (baseline) & 4/10 \\
\textbf{Real-time chunking (RTC)} & \textbf{6/10} \\
\bottomrule
\end{tabular*}
\end{table}

\subsection{Single-arm air--water transfer}
\label{sec:medium}

\begin{figure}[!t]
  \centering
  \includegraphics[width=\linewidth]{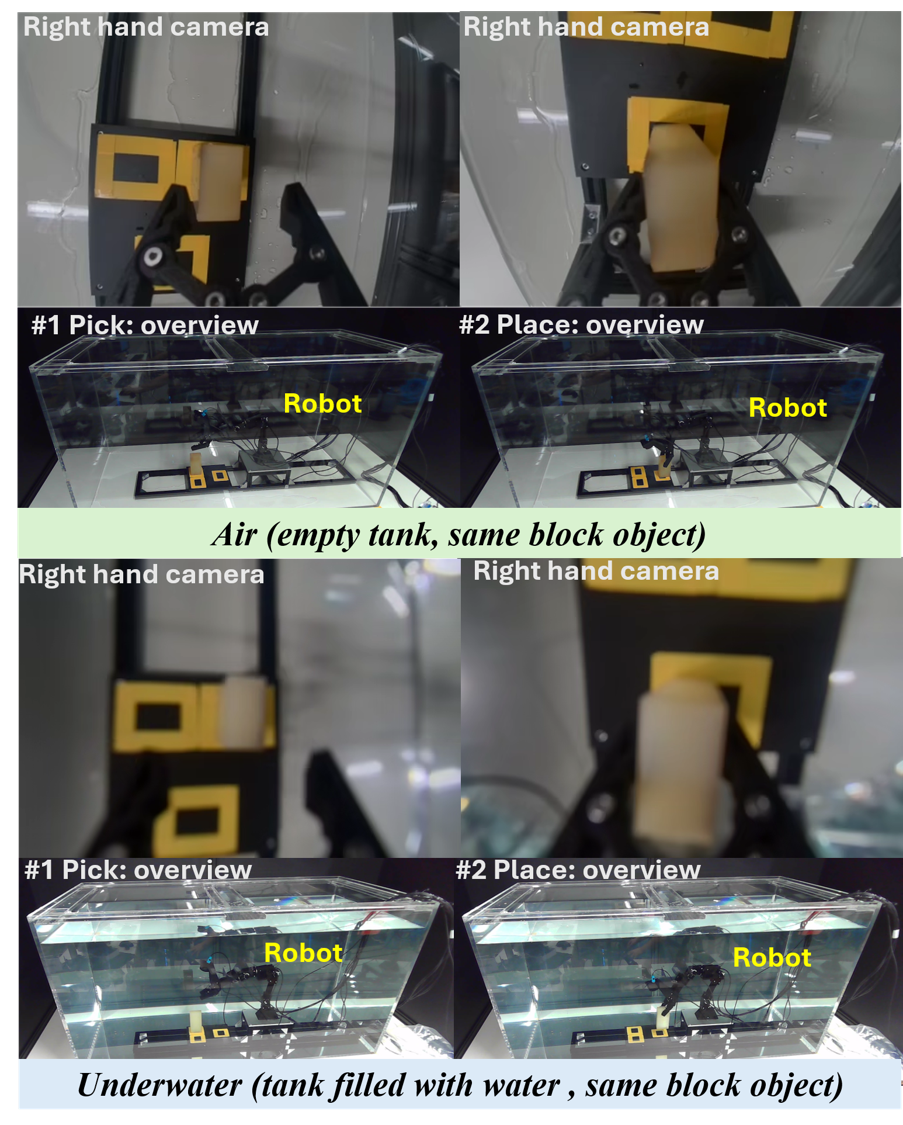}
  \caption{Single-arm pick-and-place in air (top) and underwater
    (bottom), shown from the right hand camera and an external viewpoint.
    The medium-transfer comparison uses the same task, block, and camera
    placement, with the tank empty or filled. The underwater hand-camera images appear more blurred than those captured in air, with less distinct edges around the block, gripper.}
  \label{fig:conditions}
\end{figure}

We use a separate single-arm ACT study to examine whether demonstrations
collected in one medium support execution in the other. The right follower
must grasp a block and place it fully inside the target region at rest.
Task, object, and camera placement are identical in air and underwater;
only whether the tank is filled changes (Fig.~\ref{fig:conditions}).
All policies share the robot configuration and ACT hyperparameters.
Table~\ref{tab:matrix} varies training-medium coverage while keeping the
training set at ten demonstrations per policy.

\begin{table}[t]
\caption{Single-arm air--water transfer with ACT: ten demonstrations per policy and ten evaluation trials per condition.}
\label{tab:matrix}
\centering
\tabsetup
\begin{tabular}{@{}l c c@{}}
\toprule
 & \multicolumn{2}{c}{Evaluated in} \\
\cmidrule(l){2-3}
Training episodes & underwater & air \\
\midrule
underwater 10 & \textbf{10/10} & 0/10 \\
air 10        & 0/10             & \textbf{10/10} \\
\midrule
\rowcolor{ourrow}
mixed 5+5     & \textbf{10/10} & \textbf{10/10} \\
\bottomrule
\addlinespace[2pt]
\multicolumn{3}{@{}p{0.96\columnwidth}@{}}{\scriptsize
  Mixed 5+5 denotes five demonstrations from each medium.}
\end{tabular}
\end{table}
\subsubsection{Cross-medium transfer and mixed-medium training}

Each single-medium policy achieves 10/10 in its training medium and 0/10 in
the other: 20/20 in-medium and 0/20 cross-medium in total. A single policy
trained on five demonstrations from each medium achieves 10/10 in both.
For this task, covering both media in the training set enables successful execution in both air and water without increasing the total
number of demonstrations. Filling the tank changes both perception and
physical interactions, so these results measure their combined effect.

\subsubsection{Transfer to an unseen object}

We additionally evaluate the mixed-medium policy underwater on a black rubber
block absent from training. The block differs in both color and hardness
from the white polyurethane training block. Using the same placement criterion
and 45\,s limit, the policy succeeds in 7/10 trials. This provides a limited
object-transfer result under the tested conditions; because appearance and
material properties change together, it does not isolate color generalization.

\subsection{Discussion and scope}

ULOHA supports coordinated underwater manipulation with established policy
models, while the experiments reveal sensitivity to observation conditions,
action timing, and demonstration coverage. These findings apply to the
tested setup and checkpoints. Ten trials per condition
do not establish a ranking of methods, and the experiments do not
isolate visual changes from hydrodynamic effects. Broader evaluation and
controlled comparisons remain necessary to assess generalization. ULOHA
integrates demonstration collection and policy deployment to support further
studies of underwater bimanual learning.

\section{CONCLUSIONS}
We presented ULOHA, a custom underwater bimanual platform for teleoperation,
demonstration collection, and learned-policy deployment. Nine-task experiments
demonstrate coordinated manipulation, including shared-object handling and
buoyancy-driven interception. Additional evaluations demonstrate deployment of
Diffusion Policy and SmolVLA on selected tasks.
The evaluations reveal sensitivity to bubbles, action execution horizon, and
the air--water domain shift. RTC supports deployment on release and catch,
although more trials are needed to establish its performance advantage.
Mixed-medium demonstrations enable one policy to operate in both air and
water under the tested conditions. Future work will investigate hydrodynamic
interactions, controlled visual disturbances, joint versus independent arm
policies, and force-aware teleoperation and learning.

\bibliographystyle{IEEEtran}
\bibliography{arxiv}

\end{document}